\documentclass[10pt,twocolumn,letterpaper]{article}

\usepackage{cvpr}
\usepackage{times}
\usepackage{epsfig}
\usepackage{graphicx, subcaption}
\usepackage{amsmath,mathtools}
\usepackage{amssymb}
\usepackage{multirow}
\usepackage{url}
\usepackage{amssymb}
\usepackage{pifont}
\newcommand{\cmark}{\ding{51}}%
\newcommand{\xmark}{\ding{55}}
\newcommand{\argminD}{\arg\!\min}
\usepackage{multirow}
\usepackage{array}
\usepackage{amsbsy}
\usepackage{graphicx}
\usepackage{subcaption,adjustbox}
\usepackage{soul}
\usepackage{color}
\usepackage{breqn}
\usepackage{authblk}
\usepackage{bm}

\usepackage[breaklinks=true,bookmarks=false,colorlinks]{hyperref}

\cvprfinalcopy 
\def\cvprPaperID{6322} 

\begin{document}

\title{A Kernelized Manifold Mapping to Diminish the Effect of \\ Adversarial Perturbations}


\author[1]{Saeid Asgari Taghanaki\thanks{Corresponding author}}
\author[1]{Kumar Abhishek}
\author[2]{Shekoofeh Azizi}
\author[1]{Ghassan Hamarneh}
\affil[1]{School of Computing Science, Simon Fraser University, Canada\protect \\\texttt{\{sasgarit, kabhishe, hamarneh\}@sfu.ca}}
\affil[2]{Department of Electrical and Computer Engineering, University of British Columbia, Canada\protect \\\texttt{shazizi@ece.ubc.ca}}





\maketitle

\begin{abstract}
The linear and non-flexible nature of deep convolutional models makes them vulnerable to carefully crafted adversarial perturbations. To tackle this problem, we propose a non-linear radial basis convolutional feature mapping by learning a Mahalanobis-like distance function. Our method then maps the convolutional features onto a linearly well-separated manifold, which prevents small adversarial perturbations from forcing a sample to cross the decision boundary. We test the proposed method on three publicly available image classification and segmentation datasets namely, MNIST, ISBI ISIC 2017 skin lesion segmentation, and NIH Chest X-Ray-14. We evaluate the robustness of our method to different gradient (targeted and untargeted) and non-gradient based attacks and compare it to several non-gradient masking defense strategies. Our results demonstrate that the proposed method can increase the resilience of deep convolutional neural networks to adversarial perturbations without accuracy drop on clean data.
\end{abstract}

\section{Introduction}

Deep convolutional neural networks (CNNs) are highly vulnerable to adversarial perturbations~\cite{goodfellow6572explaining,szegedy2013intriguing} produced by two main groups of attack mechanisms: white- and black-box, in which the target model parameters and architecture are accessible and hidden, respectively. In order to mitigate the effect of adversarial attacks, two categories of defense techniques have been proposed: data-level and algorithmic-level. Data-level methods include adversarial training~\cite{goodfellow6572explaining,szegedy2013intriguing}, pre-/post-processing methods (e.g., feature squeezing)~\cite{xu2017feature}, pre-processing using basis functions~\cite{shaham2018defending}, and noise removal~\cite{hendrycks2016early,meng2017magnet}. Algorithmic-level methods~\cite{folz2018adversarial,kolter2017provable,samangouei2018defense} modify the deep model or the training algorithm by reducing the magnitude of gradients~\cite{papernot2015distillation} or blocking/masking gradients~\cite{buckman2018thermometer,guo2017countering,song2017pixeldefend}. However, these approaches are not completely effective against several different white- and black-box attacks~\cite{meng2017magnet,samangouei2018defense,tramer2017ensemble} and pre-processing based methods might sacrifice accuracy to gain resilience to attacks that may never happen. Generally, most of these defense strategies cause a drop in the standard accuracy on clean data~\cite{tsipras2018there}. For more details on adversarial attacks and defenses, we refer the readers to~\cite{yuan2017adversarial}. 

Gradient masking has shown sub-optimal performance against different types of adversarial attacks~\cite{papernot2017practical,tramer2017ensemble}. Athalye et al.~\cite{athalye2018obfuscated} identified \emph{obfuscated gradients}, a special case of gradient masking that leads to a false sense of security in defenses against adversarial perturbations. They showed that 7 out of 9 recent white-box defenses relying on this phenomenon (\cite{buckman2018thermometer,dhillon2018stochastic,guo2017countering,ma2018characterizing,samangouei2018defense,song2017pixeldefend,xie2017mitigating}) are vulnerable to single step or non-gradient based attacks. They finally suggested several symptoms of defenses that rely on obfuscated gradients. Although adversarial training~\cite{madry2017towards} showed reasonable robustness against first order adversaries, this method has two major drawbacks. First, in practice the attack strategy is unknown and it is difficult to choose appropriate adversarial images for training, and second, the method requires more training time~\cite{blind_process}. Furthermore, recent studies~\cite{Towards_the_first, sharma2017breaking} show that the adversarial training overfits the $L_{\infty}$ metric while remaining highly susceptible to $L_0$, $L_1$, and $L_2$ perturbations.

As explored in the literature, successful adversarial attacks are mainly the result of models being \emph{too linear}~\cite{gilmer2018adversarial,goodfellow6572explaining, luo2015foveation, schmidt2018adversarially} in high dimensional manifolds causing the decision boundary to be close to the manifold of the training data~\cite{tanay2016boundary} and/or because of the models' \emph{low flexibility}~\cite{fawzi2015fundamental}. Another hypothesis is that adversarial examples are off the data manifold~\cite{lee2017generative, samangouei2018defense, song2017pixeldefend}. To boost the non-linearity of a model,  Goodfellow et al.~\cite{ goodfellow6572explaining} explored a variety of methods involving utilizing quadratic units and including shallow and deep radial basis function (RBF) networks. They achieved reasonably good performance against adversarial perturbations with shallow RBF networks. However, they found it difficult to train deep RBF models, leading to a high training error using stochastic gradient decent. Fawzi et al.~\cite{fawzi2018analysis} showed that support vector machines with RBF kernels can effectively resist adversarial attacks. 

Typically, a single RBF network layer takes a vector of $\mathbf{x} \in \mathbb{R}^{n}$ as input and outputs a scalar function of the input vector $f(\mathbf{x}):\mathbb{R}^{n}\rightarrow \mathbb{R}$ computed as $f\left ( \mathbf{x} \right ) = \sum_{i=1}^{P}w_{i} e^{-\beta_i D(\mathbf{x},\mathbf{c}_{i})}$, where $P$ is the number of neurons in the hidden layer, $\mathbf{c}_{i}$ is the center vector for neuron $i$, $D(\mathbf{x},\mathbf{c}_{i})$ measures the distance between $\mathbf{x}$ and $\mathbf{c}_{i}$, $w_{i}$ weights the output of neuron $i$, and $\beta_i$ corresponds to the width of the Gaussian. The input can be made linearly separable with a high probability by transforming it to a higher dimensional space (Cover's theorem~\cite{4038449})). The Gaussian basis functions, commonly used in RBF networks, are local to the center vectors, i.e., $\lim_{\left \| x \right \|\rightarrow \infty } D(\mathbf{x},\mathbf{c}_{i}) = 0$, which in  turn implies that
a small perturbation $\epsilon$ added to a sample input $\mathbf{x}$ of the neuron has an increasingly smaller effect on the output response as $\mathbf{x}$ becomes farther from the center of that neuron. Traditional RBF networks are normally trained in two sequential steps. First, an unsupervised method,  e.g.,  K-means clustering, is applied to find the RBF centers~\cite{wu2012using} and, second, a linear model with coefficients $w_{i}$ is fit to the desired outputs.

To tackle the \emph{linearity} issue of the current deep CNN models (i.e., which are persistent despite stacking several linear units and using non-linear activation functions ~\cite{gilmer2018adversarial, goodfellow6572explaining, luo2015foveation, schmidt2018adversarially}),  which results in vulnerability to adversarial perturbations, we equip CNNs with radial basis mapping kernels. Radial basis functions with Euclidean distance might not be effective as the activation of each neuron depends only on the Euclidean distance between a pattern and the neuron center. Also, since the activation function is constrained to be  symmetrical, all attributes are considered equally relevant. To address these limitations, we can add flexibility to the model by applying asymmetrical quadratic distance functions, such as the Mahalanobis distance, to the activation function in order to take into account the variability of the attributes and their correlations. However, computing the Mahalanobis distance requires complete knowledge of the attributes, which are not readily available in dynamic environments (e.g., features iteratively updated during training CNNs), and are not optimized to maximize the accuracy of the model. 

\noindent \textbf{Contributions}. In this paper, (I) we propose a new non-linear kernel based Mahalanobis distance-like feature transformation method. (II) Unlike the traditional Mahalanobis formulation, in which a constant, pre-defined covariance matrix is adopted, we propose to \emph{learn} such a ``transformation" matrix $\Psi$. (III) We propose to learn the RBF centers and the Gaussian widths in our RBF transformers. (IV) Our method adds robustness to attacks without reducing the model's performance on clean data, which is not the case for previously proposed defenses. (V) We propose a defense mechanism which can be applied to various tasks, e.g., classification, segmentation (both evaluated in this paper), and object detection.

\section{Method}
In this section, we introduce a novel asymmetrical quadratic distance function (\ref{method_dist}), and then propose a manifold mapping method using the proposed distance function (\ref{method_map}). Then we discuss the model optimization and manifold parameter learning algorithm (\ref{method_learn}).

\subsection{Adaptive kernelized distance calculation} \label{method_dist}
Given a convolutional feature map $f^{(l)} \in \mathcal{F}^{nm}$ of size $n\times m\times K^{(l)}$ for layer $l$ of a CNN, the goal is to map the features onto a new manifold $g^{(l)} \in \mathcal{G}^{nm}$ of size $n\times m\times P^{(l)}$ where classes are more likely to be linearly separable. Towards this end, we leverage an RBF transformer that takes feature vectors of $f^{(l)}_{K^{(l)}}$ as input and maps them onto a linearly separable manifold by learning a transformation matrix $\Psi^{(l)} \in\mathbb{R}^{K^{(l)}\times P^{(l)}}$ and a non-linear kernel $\kappa: \mathcal{F} \times  \mathcal{F}\rightarrow \mathbb{R}$ for which there exists a representation manifold $\mathcal{G}$ and a map $\phi: \mathcal{F} \rightarrow \mathcal{G}$ such that

\begin{equation}
    \kappa (x,z) = \phi(x).\phi(z) \ \ \ \ \  \  \forall x, z \in \mathcal{F}
\end{equation}






The $k^{th}$ RBF neuron activation function is given by

\begin{equation}
\phi^{(l)}_k = e^{-\beta^{(l)}_k D\left ( f^{(l)}_{K^{(l)}}, c^{(l)}_k \right )}
\end{equation}

\noindent where $c^{(l)}_k$ is $k^{th}$ learnable center, and the learnable parameter $\beta^{(l)}_k$ controls the width of the $k^{th}$ Gaussian function. We use a distance metric $D(.)$ inspired by the Mahalanobis distance, which refers to a distance between a convolutional feature vector and a center, and is computed as:

\begin{dmath}
D\left ( f_{K^{\left ( l \right )}}^{\left ( l \right )}, c_{k}^{\left ( l \right )} \right ) = \left ( f_{K^{\left ( l \right )}}^{\left ( l \right )}- c_{k}^{\left ( l \right )} \right )^{T} 
\left(   \Psi^{\left ( l \right )}     \right) ^{-1} \left ( f_{K^{\left ( l \right )}}^{\left ( l \right )}- c_{k}^{\left ( l \right )} \right )
\end{dmath}

\noindent where $\Psi^{(l)}$ refers to the learnable transformation matrix of size $K^{(l)}\times P^{(l)}$. To ensure that the transformation matrix $\Psi^{(l)}$ is positive semi-definite, we set $\Psi^{(l)}=AA^T$ and optimize for $A$.

Finally, the transformed feature vector $g^{(l)}(f^{(l)}_{K^{(l)}})$ is computed as:

\begin{equation}
g^{(l)}\left ( f^{(l)}_{K^{(l)}} \right ) = \sum_{k=1}^{P^{(l)}}\left ( w^{(l)}_k\cdot \phi^{(l)}_k \right )+ b^{(l)}
\end{equation}

\subsection{The proposed manifold mapping} \label{method_map}

For any intermediate layer $l$ in the network, we compute the transformed feature vector when using the RBF transformation matrix as

\begin{dmath}
g^{\left (l \right )} \left ( f^{\left (l  \right )} \right ) = \sum_{k=1}^{P^{\left ( l \right )}}\left ( w_{k}^{\left ( l \right )} \cdot \phi_{k}^{\left ( l \right )}\right ) + b^{\left ( l \right )} = \sum_{k=1}^{P^{\left ( l \right )}}\left ( w{_{k}}^{\left ( l \right )} \cdot \phi_{k}^{\left ( l \right )} \left ( f_{k}^{\left ( l \right )}\left (\theta \right ) \right )  \right) + b^{\left ( l \right )} = \sum_{k=1}^{P^{\left ( l \right )}}\left ( w_{k}^{\left ( l \right )} \right )\cdot \exp \left \{ -\beta_{k}^{\left ( l \right )} \left (f_{k}^{\left ( l \right )}\left ( \theta \right )-c_{k}^{\left ( l \right )}  \right )^{T} 
\left( \Psi^{\left ( l \right )} \right)^{-1}
\left (f_{k}^{\left ( l \right )}\left ( \theta \right )-c_{k}^{\left ( l \right )}  \right ) \right \} + b^{\left ( l \right )}
\end{dmath}

\noindent where $f^{(l)}$ is a function of the CNN parameters $\Theta$.

\begin{figure}[!ht]
\centering
  \includegraphics[width=0.5\textwidth]{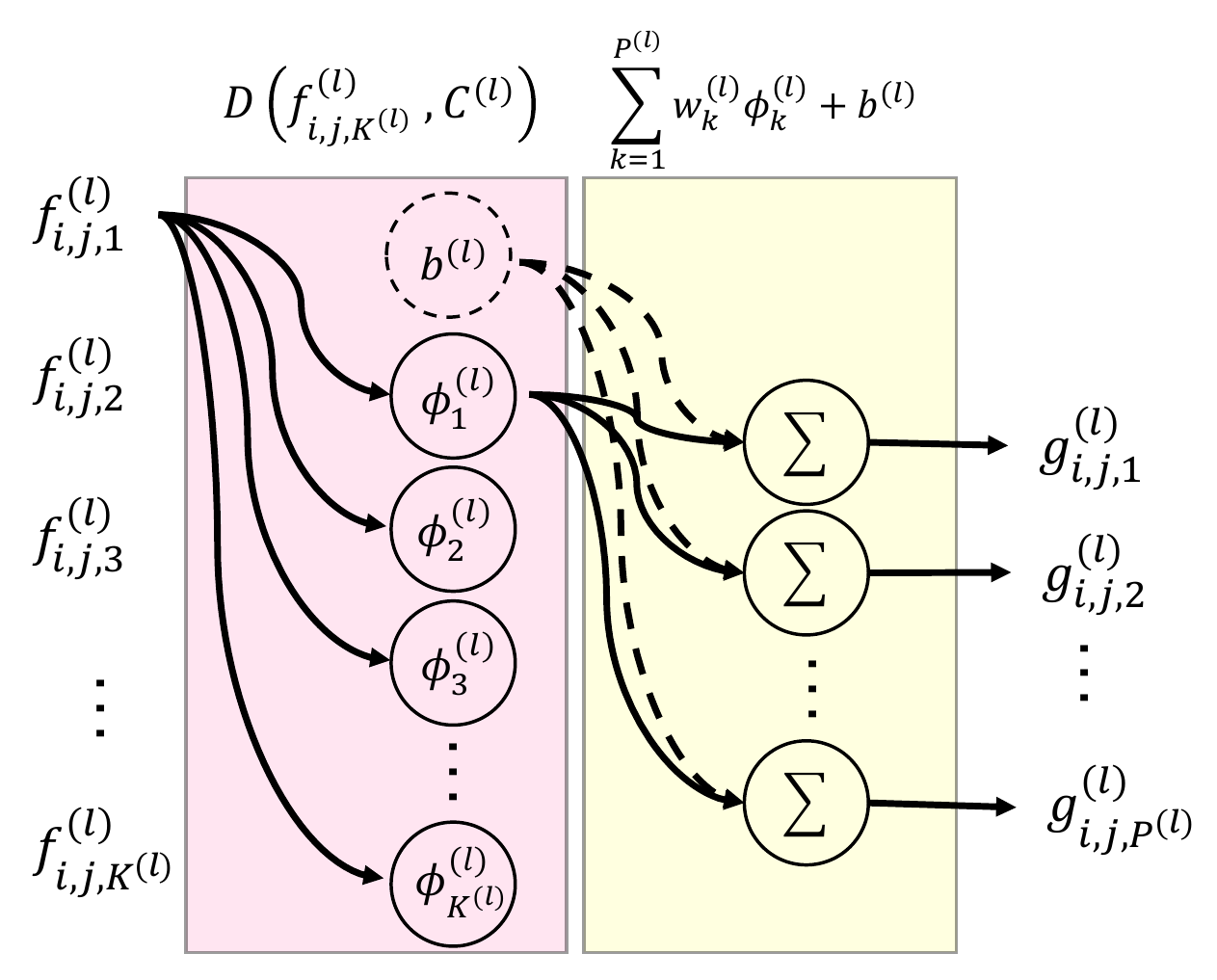}
  \caption{Detailed diagram of the proposed radial basis mapping block $g^{(l)}(f^{(l)})$. In the figure, $i=1:n$; $j=1:m$; $l=1:L$ where $n$, $m$, $K^{(l)}$, and $P^{(l)}$ are the width, the height, and the number of channels of the input and the output feature maps respectively for layer $l$.}
  \label{diag2}
\end{figure}

\begin{figure*}[!ht]
\centering
  \includegraphics[width=0.8\textwidth]{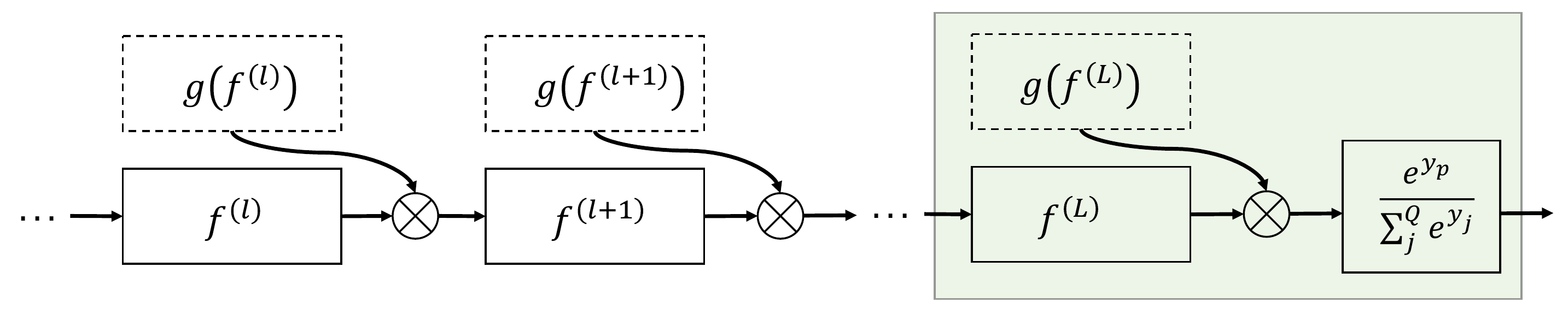}
  \caption{Placing radial basis mapping blocks in a CNN network. $\otimes$ shows concatenation operation, and $f^{(l)}$ is the output of convolution layer $l$. The output of each RBF block contains $n\in \left[1, N\right]$ channels that are concatenated to convolutional feature maps.}
  \label{diag1}
\end{figure*}

The detailed diagram of the mapping step is shown in Figure~\ref{diag2}. In each layer $l$ of network, output feature maps of the convolutional block are concatenated to the transformed feature maps (Figure~\ref{diag1}).

For the last layer (layer $L$) of the network, as shown in the green bounding box in Figure \ref{diag1}, the expression becomes

\begin{dmath}
     g^{\left ( L \right )}\left ( f^{\left (L  \right )} \right ) = \sum_{k=1}^{P^{\left ( L \right )}}\left ( w_{k}^{\left ( L \right )} \right ) \cdot \exp \left \{ -\beta_{k}^{\left ( L \right )} \left (f_{k}^{\left ( L \right )}\left ( \theta \right )- c_{k}^{\left ( L \right )}  \right )^{T} \left( \Psi^{\left ( L \right )} \right)^{-1} \left (f_{k}^{\left ( L \right )}\left ( \theta \right )-c_{k}^{\left ( L \right )}  \right ) \right \} + b^{\left ( L \right )}
\end{dmath}

The concatenation output of the last layer is given by

\begin{equation}
    y = \left[g^{(L)}(f^{(L)}), f^{(L)}\right]
\end{equation}

\subsection{Model optimization and mapping parameter learning} \label{method_learn}
All the RBF parameters, i.e., the transformation matrix $\Psi^{(l)}$, the RBF centers $C^{(l)}=\{c^{(l)}_1,c^{(l)}_2,\cdots,c^{(l)}_{N^{(l)}}\}$ (where $N^{(l)}$ denotes the number of RBF centers in layer $l$), and the widths of the Gaussians $\beta_i^{(l)}$, along with all the CNN parameters $\Theta$, are learned end-to-end using back-propagation. This approach results in the local RBF centers being adjusted optimally as they are updated based on the whole network parameters.

The categorical cross entropy loss is calculated using the softmax activation applied to the concatenation output of the last layer. The softmax activation function for the $i^{th}$ class, denoted by $\xi(y)_i$, is defined as
\begin{equation}
    \xi(y)_i = \frac{e^{y_i}}{\sum_j^Q e^{y_j}}
\end{equation}

\noindent where $Q$ represents the total number of classes. The loss is therefore defined by

\begin{equation}
    \mathcal{L} = - \sum_i^Q t_i \: log\left(\xi(y)_i\right)
\end{equation}

Since the multi-class classification labels are one-hot encoded, the expression for the loss contains only the element of the target vector $\mathbf{t}$ which is not zero, i.e., $t_p$, where $p$ denotes the positive class. This expression can be simplified by discarding the summation elements which are zero because of the target labels, and we get

\begin{equation}
    \mathcal{L} = -\:log \left(\frac{e^{y_p}}{\sum_j^Q e^{y_j}}\right)
\end{equation}
where $y_p$ represents the network output for the positive class.

Therefore, we define a loss function $\mathcal{L}$ encoding the classification error in the transformed manifold and seek $A^{*}$, $\beta^*$, $C^*$, and $\Theta^*$ that minimize $\mathcal{L}$:

\begin{equation}
A^*, C^*, \beta^*, \Theta^* = \argminD_{A,C,\beta,\Theta} \: \mathcal{L}\left ( A,C,\beta,\Theta \right ).
\end{equation}


\section{Data}
We conduct two sets of experiments: image (i) classification and (ii) segmentation. (i) For the image classification experiments, we use the MNIST~\cite{lecun1998gradient} and the NIH Chest X-Ray-14~\cite{2} (hereafter referred to as CHEST) datasets, where the latter comprises of 112,120 gray-scale images with 14 disease labels and one `no (clinical) finding' label. We treat all the disease classes as positive and formulate a binary classification task. We randomly selected 90,000 images for training: 45,000 images with ``positive" label and the remaining 45,000 with ``negative" label. The validation set comprised of 13,332 images with 6,666 images of each label. We randomly picked 200 unseen images as the test set, with 93 images of positive and 107 images of negative class, respectively. These clean (test) images are used for carrying out different adversarial attacks and the models trained on clean images are evaluated against them. (ii) For the image segmentation task experiments, we use the 2D RGB skin lesion dataset from the 2017 IEEE ISBI International Skin Imaging Collaboration (ISIC) Challenge~\cite{codella2017skin} (hereafter referred to as SKIN). We trained on a set of 2,000 images and test on an unseen set of 150 images.

\section{Experiments and results}
In this section, we report the results of several experiments for two tasks of classification and segmentation. We first start with MNIST as it has extensively been used for evaluating adversarial attacks and defenses. Next, we show how the proposed method is applicable to another classification dataset and segmentation task.

\subsection{Evaluation on classification task}
In this section, we analyze the performance of the proposed method on two different classifications datasets MNIST and CHEST. In Table~\ref{tablemnist}, we report the results of the proposed method on MNIST dataset when attacked by different targeted and un-targeted attacks i.e., fast gradient sign method (FGSM)~\cite{kurakin2016adversarial}, basic iterative method (BIM)~\cite{kurakin2016adversarial}, projected gradient descent (PGD)~\cite{madry2017towards}, Carlini \& Wagner method (C\&W)~\cite{carlini2017towards}, and momentum iterative method (MIM)~\cite{dong2018boosting} (the winner of NIPS 2017 adversarial attacks competition). The proposed method (i.e., PROP) successfully resists all the attacks (with both $L_{\infty}$ and $L_{2}$ perturbations) for which the 3-layers CNN (i.e., ORIG) network almost completely fails e.g., for the strongest attack (i.e., MIM) the proposed method achieves $64.25\%$ accuracy while the original CNN network obtains almost zero ($0.58\%$) accuracy. Further, we test the proposed method with two non-gradient based attacks: simultaneous perturbation stochastic approximation (SPSA)~\cite{uesato2018adversarial} and Gaussian additive noise (GN)~\cite{rauber2017foolbox} to show that the robustness of the proposed method is not because of gradient masking. We compare our results to other defenses e.g., Binary CNN, Nearest Neighbour (NN) model, Analysis by Synthesis (ABS), Binary ABS~\cite{Towards_the_first} and Fortified Networks~\cite{lamb2018fortified}. Looking at the Binary ABS and ABS results in Table~\ref{tablemnist}, the former generally outperforms the latter, but it should be noted that Binary ABS is applicable only to simple datasets, e.g., MNIST, as it leverages binarization. Although Fortified Net outperforms our method for the FGSM attack, it has been tested only on gradient-based attacks, and therefore, it is unclear how it would perform against gradient-free attacks such as SPSA and GN.

\begin{table*}[h!]
\centering
\caption{Classification accuracy under different attacks tested on MNIST dataset. FGSM: $\epsilon = 0.3$; BIM: $\epsilon = 0.3$ and iterations = 5; MIM: $\epsilon = 0.3$, iterations = 10, and decay factor = 1; PGD: $\epsilon = 0.1$, iterations = 40; C\&W: iterations = 50, GN: $\epsilon = 20$; SPSA: $\epsilon = 0.3$. ``n/a" denotes that the corresponding entry was not reported in the respective paper.}
\begin{tabular}{l|c|cc|ccccc}
\hline
\multirow{2}{*}{Models} & \multirow{2}{*}{Clean} & \multicolumn{2}{c|}{$L{_2}$} & \multicolumn{5}{c}{$L{_\infty}$} \\ \cline{3-9} 
 &  & C\&W~\cite{carlini2017towards} & GN~\cite{rauber2017foolbox} & FGSM~\cite{kurakin2016adversarial} & BIM~\cite{kurakin2016adversarial} & MIM~\cite{dong2018boosting} & PGD~\cite{madry2017towards} & SPSA~\cite{uesato2018adversarial} \\ \hline
ORIG~\cite{papernot2018cleverhans} & 0.9930 & 0.1808 & 0.7227 & 0.0968 & 0.0070 & 0.0051 & 0.1365 & 0.3200 \\
Binary CNN~\cite{Towards_the_first} & 0.9850 & n/a & 0.9200 & 0.7100 & 0.7000 & 0.7000 & n/a  & n/a \\
NN~\cite{Towards_the_first} & 0.9690 & n/a & 0.9100 & 0.6800 & 0.4300 & 0.2600 & n/a & n/a \\
Binary ABS~\cite{Towards_the_first} & 0.9900 & n/a & 0.8900 & 0.8500 & \textbf{0.8600} & \textbf{0.8500} & n/a & n/a \\
ABS~\cite{Towards_the_first} & 0.9900 & n/a & \textbf{0.9800} & 0.3400 & 0.1300 & 0.1700 & n/a & n/a \\
Fortified Net~\cite{lamb2018fortified} & 0.9893 & 0.6058 &n/a & \textbf{0.9131} & n/a & n/a & 0.7954 & n/a \\
PROP & \textbf{0.9942} & \textbf{0.9879} & 0.7506 & 0.8582 & 0.7887 & 0.6425 & \textbf{0.8157} & \textbf{0.7092} \\ \hline
\end{tabular}
\label{tablemnist}
\end{table*}

\begin{figure}
\centering
\begin{adjustbox}{center}
\begin{subfigure}[b]{\columnwidth}
    \centering
    \includegraphics[width=\columnwidth]{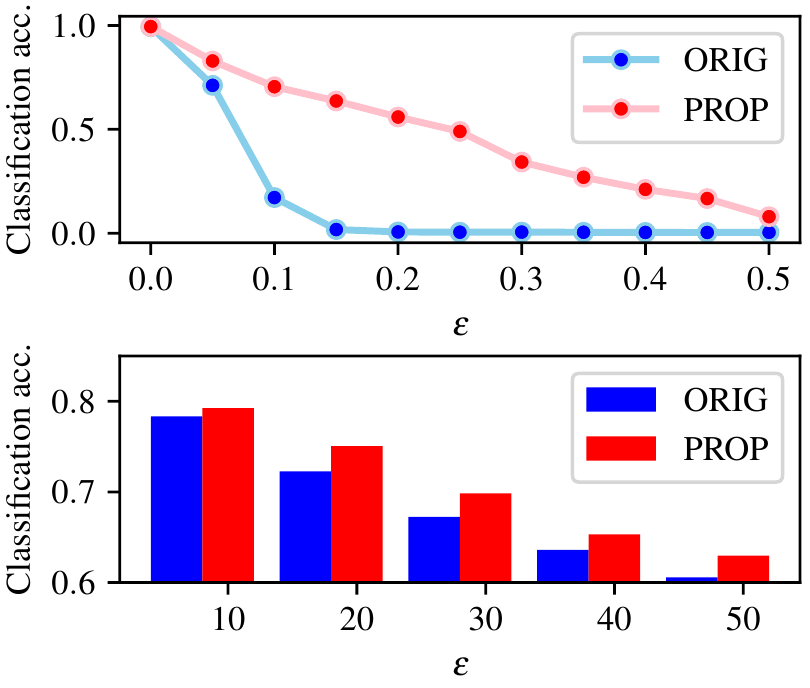}
    \caption{top: PGD and bottom: GN attack \\~\\}
    \label{fig:1}
\end{subfigure}
\end{adjustbox}
\begin{adjustbox}{center}
\begin{subfigure}[b]{\columnwidth}
    \centering
    \includegraphics[width=\textwidth]{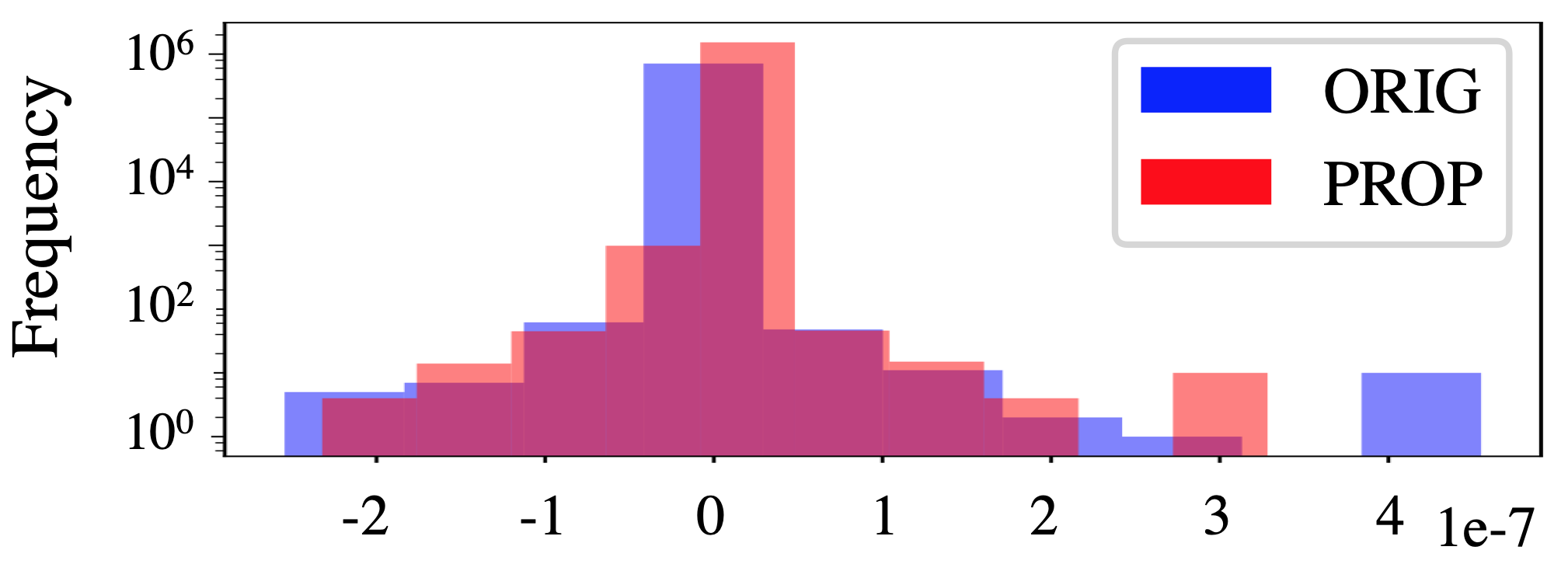}
    \caption{Gradient distribution}
    \label{fig:2}
\end{subfigure}
\end{adjustbox}
\caption{Gradient masking analysis: (a) accuracy-distortion plots and (b) gradient distribution for the proposed method and the original CNN}
\label{gn}
\end{figure}

 To ensure that the robustness of the proposed method is not due to masked/obfuscated gradient, as suggested by~\cite{athalye2018obfuscated}, we test the proposed feature mapping method based on several characteristic behaviors of defenses that cause obfuscated gradients to occur. a) As reported in Table~\ref{tablemnist}, one-step attacks (e.g., FGSM) did not perform better than iterative attacks (e.g., BIM, MIM); b) According to Tables~\ref{table1} and~\ref{table2}, black-box attacks did not perform better than white-box ones; c) as shown in Figure~\ref{gn} (a and b), larger distortion factors monotonically increase the attack success rate; d) the proposed method performs well against gradient-free attacks e.g., GN and SPSA. The subplot (b) in Figure~\ref{gn} also indicates that the robustness of the proposed method is not because of numerical instability of gradients.

Next, to quantify the compactness and separability of different clusters/classes, we evaluate the features produced ORIG and PROP methods with clustering evaluation techniques such as mutual information based score~\cite{vinh2010information}, homogeneity and completeness~\cite{rosenberg2007v}, Silhouette coefficient ~\cite{rousseeuw1987silhouettes}, and Calinski-Harabaz index~\cite{calinski1974dendrite}. Both Silhouette coefficient and Calinski-Harabaz index quantify how well clusters are separated from each other and how compact they are without taking into account the ground truth labels, while mutual information based score, homogeneity, and completeness scores evaluate clusters based on labels. As reported in Table~\ref{table4}, when the original CNN network applies radial basis feature mapping it achieves considerably higher scores (for all the metrics, higher values are better). As both the original and the proposed method achieved high classification test accuracy i.e., $\sim99\%$, the difference in scores for label based metrics, i.e., mutual information based, homogeneity, and completeness, scores are small.  

\begin{table}[h!]
\setlength{\tabcolsep}{4pt}
\centering
\caption{Feature mapping analysis via intra-class compactness and inter-class separability measures of the MNIST dataset for original 3-layer CNN versus the proposed method. The abbreviated column headers are Silhouette, Calinski, Mutual Information, Homogeneity, and Completeness metrics, respectively.}
\begin{tabular}{lccccc}
\hline
      & Sil. & Cal. & MI & Homo. & Comp.  \\ \hline
ORIG~\cite{papernot2018cleverhans} & 0.2612     &1658.20    &0.9695             & 0.9696      & 0.9721   \\
PROP  & \textbf{0.4284}     & \textbf{2570.42}        & \textbf{0.9720}             & \textbf{0.9721}     & \textbf{0.9815}        \\ \hline
\end{tabular}
\label{table4}
\end{table}

\begin{figure*}[h!]
  \includegraphics[width=\linewidth]{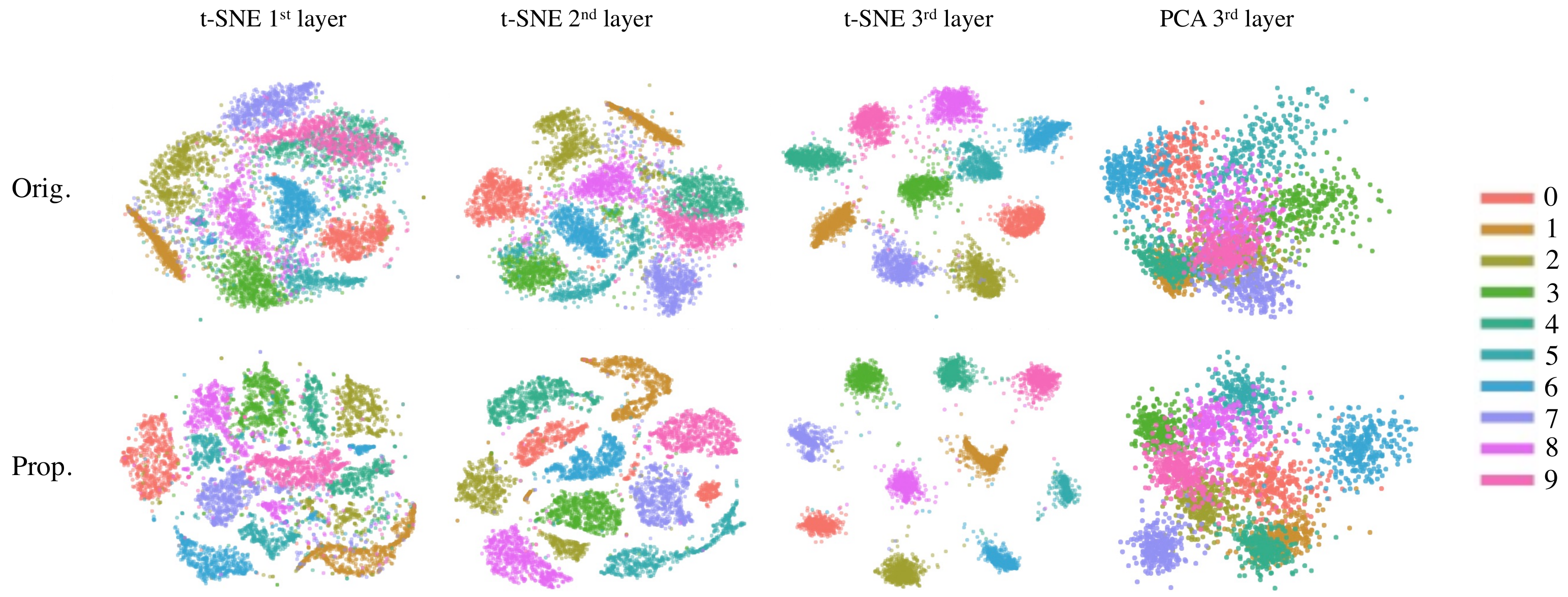}
  \caption{Feature space visualization at different layers via t-SNE and PCA of MNIST dataset produced via  original 3-layer CNN network and the proposed method.}
  \label{tsne}
\end{figure*}

In Figure~\ref{tsne}, we visualize the feature spaces of each layer in a simple 3-layer CNN using t-SNE~\cite{maaten2008visualizing} and PCA~\cite{jolliffe2011principal} methods by reducing the high dimensional feature space into two dimensions. As can be seen, the proposed radial basis feature mapping reduces intra-class and increases inter-class distances. 

To test the robustness of the proposed method on CHEST, we follow the strategy of Taghanaki et al.~\cite{taghanaki2018vulnerability}. We select Inception-ResNet-v2~\cite{szegedy2017inception} and modify it by the proposed radial basis function blocks. According to the study done by Taghanaki et al.~\cite{taghanaki2018vulnerability}, we focus on the most effective attacks in term of imperceptibility and power i.e., gradient-based attacks (basic iterative method~\cite{kurakin2016adversarial}: BIM and L1-BIM). We also compare the proposed method with two defense strategies: Gaussian data augmentation (GDA)~\cite{zantedeschi2017efficient} and feature squeezing (FSM)~\cite{xu2017feature}. GDA is a data augmentation technique that augments a dataset with copies of the original samples to which Gaussian noise has been added. FSM method reduces the precision of the components of the original input by encoding them with a smaller number of bits. In Table~\ref{table5}, we report the classification accuracy of different attacks and defenses (including PROP) on CHEST.

\begin{table}[h!]
\setlength{\tabcolsep}{4pt}
\centering
\caption{Classification accuracy on CHEST for different attacks and defenses.}
\begin{tabular}{lcccccc}
\hline
~                  & \multicolumn{5}{c}{Defense}\\
\hline
Attack                  & Iteration & ORIG       & GDA    & FSM     & PROP \\ \hline
\multirow{1}{*}{$L{_1}$ \text{BIM}}~\cite{kurakin2016adversarial} & 5         & 0         & 0      & 0.55    & \textbf{0.63} \\
\multirow{1}{*}{$L{_\infty}$ \text{BIM}}~\cite{kurakin2016adversarial}    & 5         & 0         & 0      & 0.54    & \textbf{0.65}  \\
\multirow{1}{*}{Clean}  & -         & 0.74      & \textbf{0.75}    & 0.57   & 0.74   \\\hline
\end{tabular}
\label{table5}
\end{table}

\subsection{Evaluation on segmentation task}
To assess the segmentation vulnerability to adversarial attacks, we apply the dense adversary generation (DAG) method proposed by Xie et al.~\cite{ciahng2017} to two state-of-the-art segmentation networks: U-Net~\cite{ronneberger2015u} and V-Net~\cite{milletari2016v} under both white- and black-box conditions. We compare the proposed feature mapping method to other defense strategies e.g., Gaussian and median feature squeezing~\cite{xu2017feature} (FSG and FSM, respectively) and adversarial training~\cite{goodfellow6572explaining} (ADVT) on SKIN. From Table~\ref{table1}, it can be seen that the proposed method is more robust to adversarial attacks and when applied to U-Net, its performance deteriorates much lesser than the next best method (ADVT) 1.60\% and 6.39\% vs 9.44\% and 13.47\% after 10 and 30 iterations of the attack with $\gamma = 0.03$. Similar performance was also observed for V-Net, where the accuracy drop of the proposed method using the same $\gamma$ for 10 iterations was 8.50\%, while the next best method (ADVT) dropped by 11.76\%. It should also be noted that applying the feature mapping led to an improvement in the segmentation accuracy on clean (non-attacked/unperturbed) images, and the performance increased to 0.7780 from the original 0.7743 for U-Net, and to 0.8213 from the original 0.8070 for V-Net.

\begin{figure*}[h!]
  \includegraphics[width=\textwidth]{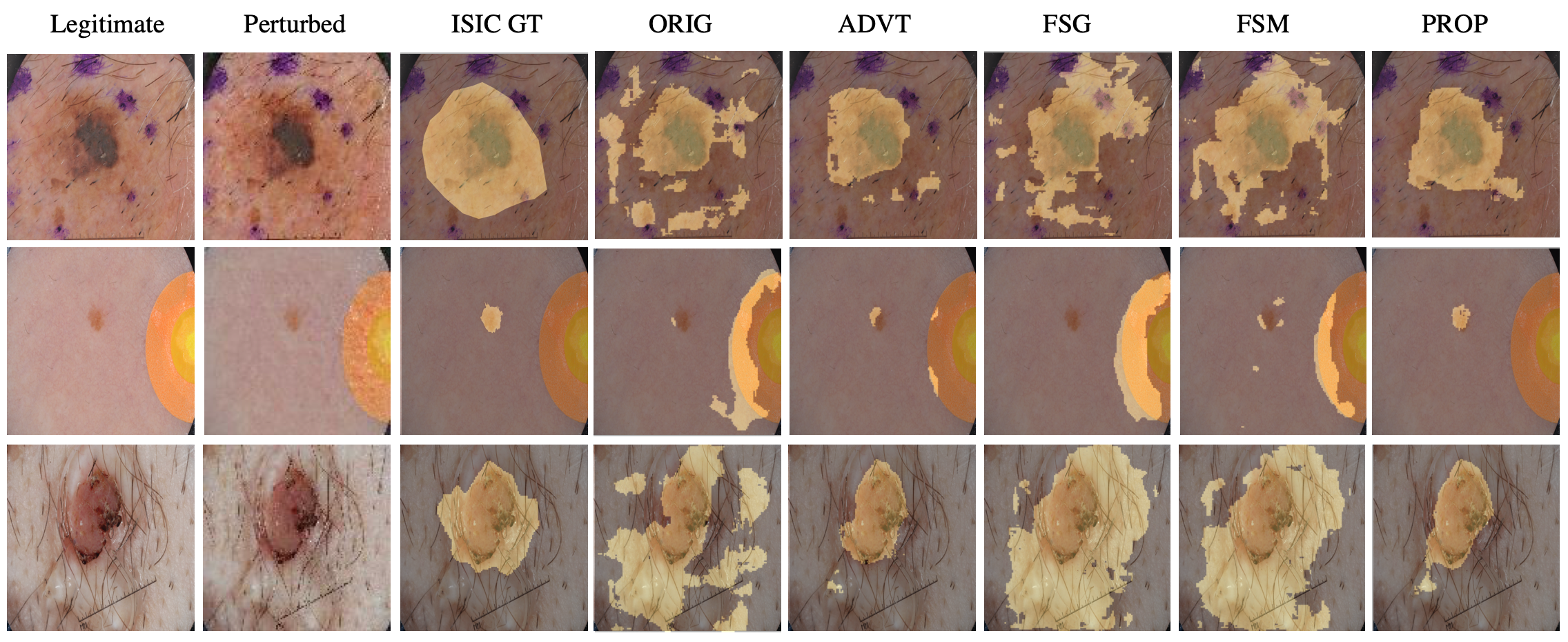}
  \caption{Lesion segmentation sample results on SKIN for different defense strategies applied with U-Net.}
  \label{skin_qual}
\end{figure*}

Figure~\ref{skin_qual} visualizes the segmentation results of a few samples from SKIN for different defense strategies. As shown, the proposed method obtains  the closest results to the ISIC ground truth (GT) than all other methods. Although adversarial training (ADVT) also produces promising segmentation results, it requires knowledge of the adversary in order to perform robust optimization which is almost impossible to obtain in practice since the attack is unknown. However, the proposed method does not have such a dependency.

\begin{table*}[h!]
\centering
\caption{Segmentation results (average DICE $\pm$ standard error) of different defense mechanisms compared to the proposed radial basis feature mapping method for V-Net and U-Net under DAG attack. $10i$ and $30i$ refer to 10 and 30 iterations of attack, respectively.}

\begin{tabular}{llccc}
\hline
Network                & Method & Clean               & $10i$ (\% Accuracy drop)          & $30i$ (\% Accuracy drop)         \\ \hline
\multirow{5}{*}{U-Net~\cite{ronneberger2015u}} & ORIG~\cite{ronneberger2015u}    & $0.7743 \pm 0.0202$ & $0.5594 \pm 0.0196 (27.75\%)$  & $0.4396 \pm 0.0222 (43.23\%)$ \\
                       & FSG~\cite{xu2017feature}    & $0.7292 \pm 0.0229$ & $0.6382 \pm 0.0206 (15.58\%)$  & $0.5858 \pm 0.0218 (24.34\%)$ \\
                       & FSM~\cite{xu2017feature}    & $0.7695 \pm 0.0198$ & $0.6039 \pm 0.0199 (22.01\%)$  & $0.5396 \pm 0.0211 (30.31\%)$ \\
                       & ADVT~\cite{goodfellow6572explaining}   & $0.6703 \pm 0.0273$ & $0.7012 \pm 0.0255 (9.44\%)$   & $0.6700 \pm 0.0260 (13.47\%)$ \\
                       & PROP   & \textbf{0.7780} $\pm$ \textbf{0.0209} & \textbf{0.7619 }$\pm$ \textbf{0.0208} (\textbf{1.60}\%)   & \textbf{0.7248} $\pm$ \textbf{0.0226} (\textbf{6.39}\%)  \\
                       &        &                     &                               &                              \\
\multirow{5}{*}{V-Net~\cite{milletari2016v}} & ORIG~\cite{ronneberger2015u}    & $0.8070 \pm 0.0189$ & $0.5320 \pm 0.0207 (34.10\%)$  & $0.3865 \pm 0.0217 (52.10\%)$ \\
                       & FSG~\cite{xu2017feature}    & $0.7886 \pm 0.0205$ & $0.6990 \pm 0.0189 (13.38\%)$  & $0.6840 \pm 0.0188 (15.24\%)$ \\
                       & FSM~\cite{xu2017feature}   & $0.8084 \pm 0.0189$ & $0.5928 \pm 0.0209 (26.54\%)$  & $0.5144 \pm 0.0218 (36.26\%)$ \\
                       & ADVT~\cite{goodfellow6572explaining}   & $0.7924 \pm 0.0162$ & $0.7121 \pm 0.0174 (11.76\%)$  & \textbf{0.7113} $\pm$ \textbf{0.0179} (\textbf{11.85}\%) \\
                       & PROP   & \textbf{0.8213} $\pm$ \textbf{0.0177} & \textbf{0.7384} $\pm$ \textbf{0.0169} (\textbf{8.50}\%)   & $0.6944 \pm 0.0178 (13.95\%)$ \\ \hline
\label{table1}
\end{tabular}
\end{table*}

As reported in Table~\ref{table2}, under black-box attack, the proposed method is the best performing method across all 12 experiments except for one in which the accuracy of the best method was just $0.0022$ higher (i.e., $0.7284 \pm 0.2682$ vs $0.7262 \pm 0.2621$). However, it should be noted that the standard deviation of the winner is larger than the proposed method.

\begin{table*}[h!]
\centering
\caption{Segmentation DICE $\pm$ standard error scores of black-box attacks; adversarial images were produced with methods in first left column and tested with methods in the first row. U-PROP and V-PROP refer to equipped U-Net and V-Net with our mapping method.}
\begin{tabular}{lcccc}
\hline
- & U-Net~\cite{ronneberger2015u}                    & U-PROP                   & V-Net~\cite{milletari2016v}                    & V-PROP                   \\
U-Net~\cite{ronneberger2015u}                    & - & \textbf{0.7341} $\pm$ \textbf{0.0205}      & $0.6364 \pm 0.0189$      & $0.7210 \pm 0.0189$      \\
U-PROP                   & \textbf{0.7284} $\pm$ \textbf{0.0219}     & - & $0.6590 \pm 0.0218$      & $0.7262 \pm 0.0241$      \\
V-Net~\cite{milletari2016v}                    & $0.7649 \pm 0.0168$      & \textbf{0.7773} $\pm$ \textbf{0.0167}    & - & $0.7478 \pm 0.2090$      \\
V-PROP                   & $0.7922 \pm 0.0188$      & \textbf{0.7964} $\pm$ \textbf{0.0192}     & $0.6948 \pm 0.0171$      & - \\ \hline
\end{tabular}
\label{table2}
\end{table*}

Next, we analyze the usefulness of learning the transformation matrix ($\Psi$) and width of the Gaussian ($\beta$) in our Mahalanobis-like distance calculation. As can be seen in Table~\ref{table3}, in all the cases, i.e., testing with clean images and images 10 and 30 iterations of attack, our method with $\Psi$ and $\beta$ achieved higher performance.

\begin{table*}[h!]
\centering
\caption{Ablation study over the usefulness of learning the transformation matrix $\Psi$ and $\beta$ on SKIN for V-Net; mean $\pm$ standard error}
\begin{tabular}{lccccc}
\hline
                        & \multicolumn{1}{c}{$\Psi$} & \multicolumn{1}{c}{$\beta$} & \multicolumn{1}{c}{Dice} & \multicolumn{1}{c}{FPR} & \multicolumn{1}{c}{FNR} \\ \hline
\multirow{4}{*}{Clean}  & \xmark                         & \cmark                      & $0.7721 \pm 0.0210$        & $0.0149 \pm 0.0022$       & $0.2041 \pm 0.0245$       \\
                        & \cmark                         & \xmark                      & $0.8200 \pm 0.0163$        & $0.0177 \pm 0.0026$       & $0.1547 \pm 0.0188$       \\
                        & \xmark                         & \xmark                      & $0.8002 \pm 0.0184$        & $0.0137 \pm 0.0022$       & $0.1883 \pm 0.0211$       \\
                        & \cmark                         & \cmark                      & \textbf{0.8213} $\pm$ \textbf{0.0177}        & $0.0141 \pm 0.0020$       & $0.1706 \pm 0.0200$      \\
                        &                                &                             &                          &                         &                         \\
\multirow{4}{*}{$10 i$} & \xmark                         & \cmark                      & $0.6471 \pm 0.0213$        & $0.0437 \pm 0.0052$       & $0.1992 \pm 0.0261$       \\
                        & \cmark                         & \xmark                      & $0.7010 \pm 0.0161$        & $0.0606 \pm 0.0054$       & $0.1020 \pm 0.0166$       \\
                        & \xmark                         & \xmark                      & $0.6740 \pm 0.0187$        & $0.0458 \pm 0.0037$       & $0.1472 \pm 0.0217$       \\
                        & \cmark                         & \cmark                      & \textbf{0.7384} $\pm$ \textbf{0.0169}        & $0.0444 \pm 0.0041$       & $0.1234 \pm 0.0186$       \\
                        &                                &                             &                          &                         &                         \\
\multirow{4}{*}{$30 i$} & \xmark                         & \cmark                      & $0.6010 \pm 0.0221$        & $0.0371 \pm 0.0030$      & $0.2304 \pm 0.0273$       \\
                        & \cmark                         & \xmark                      & $0.6458 \pm 0.0180$       & $0.0633 \pm 0.0042$       & $0.1164 \pm 0.0183$       \\
                        & \xmark                         & \xmark                      & $0.6188 \pm 0.0188$        & $0.0615 \pm 0.0040$       & $0.1384 \pm 0.0205$       \\
                        & \cmark                         & \cmark                      & \textbf{0.6944} $\pm$ \textbf{0.0179}        & $0.0418 \pm 0.0038$       & $0.1489 \pm 0.0205$       \\ \hline
\end{tabular}
\label{table3}
\end{table*}

\begin{figure}[h!]
   \centering
   \includegraphics[width=\columnwidth]{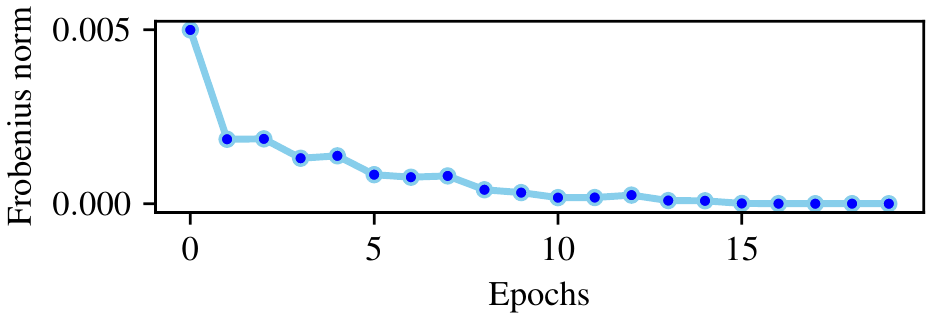}
   \caption{Convergence: Plotting the Frobenius norm of the change in $\Psi$ between two consecutive epochs.}
   \label{norm}
 \end{figure}

Figure~\ref{norm} shows how the $\Psi$ of a single layer converges after a few epochs on the MNIST dataset. The Y-axis is the Frobenius norm of the change in $\Psi$ between two consecutive epochs. Adopting the value $\Psi$ converges to results in a superior performance (e.g., mean Dice 0.8213 vs 0.7721, as reported in Table~\ref{table3}) compared to when we do not optimize for $\Psi$, i.e., hold it constant (identity matrix).

\section{Implementation details}
\subsection{Classification experiments} 

$\bullet$ MNIST experiments: For both the original CNN with 3-layers (i.e., ORIG) and the proposed method (i.e., PROP), we used a batch size of 128 and Adam optimizer with learning rate of 0.001. 

$\bullet$ CHEST: Inception-ResNet-v2 network was trained with a batch size of 4 with RMSProp optimizer \cite{16} with a decay of 0.9 and $\epsilon = 1$ and an initial learning rate of 0.045, decayed every 2 epochs using an exponential rate of 0.94. 
    
 For all the gradient based attacks applied in the classification part, we used the CleverHans library~\cite{papernot2018cleverhans}, and for the Gaussian additive noise attack, we used FoolBox~\cite{rauber2017foolbox}.

\subsection{Segmentation experiments}

For both U-Net and V-Net, we used a batch size of 16, ADADELTA optimizer with learning rate of 1.0, rho = 0.95, and decay = 0.0. We tested the DAG method with 10 and 30 iterations and perturbation factor $\gamma = 0.03$. For FSM and FSG defenses, we applied a window size of $3\times 3$ and a standard deviation of 1.0, respectively.

\section{Conclusion}
We proposed a nonlinear radial basis feature mapping method to transform layer-wise convolutional features into a new manifold, where improved class separation reduces the effectiveness of perturbations when attempting to fool the model. We evaluated the model under white- and black-box attacks for two different tasks of image classification and segmentation and compared our method to other non-gradient based defenses. We also performed several tests to ensure that the robustness of the proposed method is neither because of numerical instability of gradients nor because of gradient masking. In contrast to previous methods, our proposed feature mapping improved the classification and segmentation accuracy on both clean and perturbed images.

\section*{Acknowledgement}
Partial funding for this project is provided by the Natural Sciences and Engineering Research Council of Canada (NSERC). The authors are grateful to the NVIDIA Corporation for donating a Titan X GPU used in this research.

{\small
\bibliographystyle{ieee}
\bibliography{egpaper_final}
}



\end{document}